# Exact Topology Reconstruction of Radial Dynamical Systems with Applications to Distribution System of the Power Grid

Saurav Talukdar[1], Deepjyoti Deka[2], Donatello Materassi[3] and Murti Salapaka[4]

*Abstract*—In this article we present a method to reconstruct the interconnectedness of dynamically related stochastic processes, where the interactions are bi-directional and the underlying topology is a tree. Our approach is based on multivariate Wiener filtering which recovers spurious edges apart from the true edges in the topology reconstruction. The main contribution of this work is to show that all spurious links obtained using Wiener filtering can be eliminated if the underlying topology is a tree based on which we present a three stage network reconstruction procedure for trees. We illustrate the effectiveness of the method developed by applying it on a typical distribution system of the electric grid.

## I. INTRODUCTION

Networks underpin a powerful framework for modeling and analysis of large scale dynamical systems. Applications include neuroscience [1], financial markets [2], protein dynamics [3], climate sciences [4] and the power grid [5]. Moreover, networks play an indispensable role in building foundational aspects of control theory [6], statistical inference [7] and optimization theory[8]. The compactness of representation and the capability of unveiling influences, cause-effect relationships and dependencies amongst many variables are some of the key attributes enabled by network based approaches [9], [10]. An essential aspect of many studies is to determine a graphical representation of how multiple sub-systems/agents interact from measured time series data. It is often the case that active manipulation of the system is prohibited or not possible; for example, in financial markets the prices of stocks are available as data but it is not possible (or not allowed) to manipulate the prices. In many cases, the influences between sub-systems/ agents is mutual, thus separating source and destination or cause and effect in such cases is not meaningful.

In this article, we are concerned with the task of unveiling the network topology that relates multiple linear dynamical systems from temporal data, where it is not possible to excite the system externally. Here, we assume that the underlying network is bi-directed, that is, the influences between agents is mutual and describing cause-effect relationships is not obvious. We further restrict the study to systems where the interaction flow is well characterized by a tree structure.

[1]Saurav Talukdar is with Department of Mechanical Engineering, University of Minnesota, Minneapolis, USA, sauravtalukdar@umn.edu
[2]Deepjyoti Deka is with Los Alamos National Lab, Los Alamos, USA, deepjyoti@lanl.gov
[3]Donatello Materassi is with Department of Electrical and Computer Engineering, University of Tennessee, Knoxville, USA, dmateras@utk.edu
[4]Murti V. Salapaka is with Department of Electrical and Computer Engineering, Minneapolis, USA, murtis@umn.edu

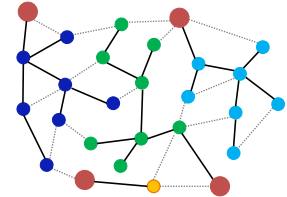

Fig. 1. Distribution network with roots represented by large red nodes. The operational edges are formed by solid lines (black). Dotted grey lines represent open switches. Non-root nodes within each tree are marked with the same color [18].

Learning the structure of a network of static random variables is an active research area in many fields since almost half a century [10], [9], [11], [12], [13]. Recently, there has been growing interest on determining the structure of a network of dynamically related systems. Network reconstruction for a collection of wide sense stationary processes related by linear dynamical systems is approached using inverse of the power spectral density matrix in [14], Wiener filtering in [15], [16] and mutual information in [17]. Our work builds on [15], where it is shown that multivariate Wiener filtering recovers the Markov blanket of each node in the network which includes spurious edges in addition to the true edges in the underlying network. The main contribution of this article is a method that eliminates the spurious links to obtain an exact reconstruction of the underlying topology for dynamical systems that have bi-directional interactions with the interaction topology described by a tree. We instantiate the motivation and the results to a power grid. The theory and results are inspired by [18] where aspects of the power grid are modeled in a static framework.

The power grid is a large engineered infrastructural network that has facilitated uninterrupted availability of energy fueling huge technological advances. Power distribution systems, which consist of the medium and low-voltage lines that connect the distribution substations to the end-users are structurally characterized by interconnections that can be modeled via a radial topology [19]. The radial topology may be altered over time by changing the operational lines selected from a set of permissible lines, while maintaining the tree structure [20] (see Fig. 1). In recent years, the proliferation of smart controllable devices and household generators has led to greater focus on the estimation and control of the distribution system. A critical need, here, is the accurate estimation of the current radial topology that may change over time due to unreported maintenance or faults. The estimation problem is further exacerbated by the historical low-penetration of real-time line meters in the distribution grid. The challenge is being partly met by the use of advanced meters like PMUs (Phase Measurement Units) [21] and distribution micro-PMUs [22] that provide high fidelity measurements of the state of the buses. Prior research directions in this area include learning using inverse covariance matrices [23], trends in voltage variance [20], [24], [25], graphical model learning [18] and maximum likelihood schemes [26]. However, the above indicated studies assume the measured data to be independent samples from a static distribution. Such an assumption may not be accurate as fast sampled (sub-sec sampling) data in the grid arise from the evolution of states that are not static but dynamic. Here, we avoid the static assumption and consider the nodal measurements to arise from the swing dynamics in the power grid [5]. Under the dynamical framework, we apply our analytical development pertinent to bi-directed dynamically related processes, to provably learn the correct radial topology of the grid. We demonstrate the performance of our learning framework on IEEE test dynamic networks.

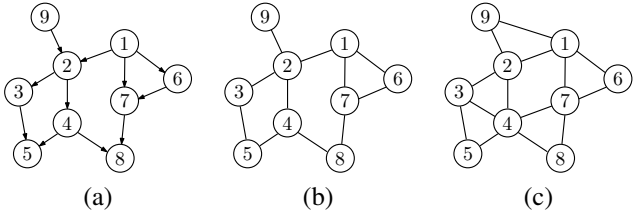

Fig. 2. (a) A directed graph, (b) its topology (nodes 2 and 3 are neighbors, 2 and 5 are two hop neighbors) and (c) its kin graph.

In the next section we introduce notions from graph theory, which are utilized later, following which we introduce the framework of Linear Dynamic Graphs in Section III. In Section IV the Wiener filtering based network topology reconstruction algorithm is discussed. Then we present results and algorithms to obtain an exact reconstruction for a tree topology in Section V, after which we introduce the swing dynamics of a network of generators in Section VI to illustrate the algorithms presented with examples in Section VII. We end with conclusions in Section VIII.

## II. PRELIMINARIES

In this section, basic notions of graph theory that are useful for the subsequent development are recalled [27].

*Definition 1 (Directed and Undirected Graphs):* An undirected graph $G$ is a pair $(V, A)$ where $V$ is a set of vertices or nodes and $A$ is a set of edges or arcs, which are unordered subsets $(i, j), i, j \in V$. We refer to $j$ as the neighbor of $i$ and vice versa. If arcs in $A$ are ordered, it is called a directed (or oriented) graph.

*Definition 2 (Topology of a graph):* Given an oriented graph $G = (V, A)$, its topology $top(G)$ is defined as the undirected graph $G' = (V, A')$ that is obtained by removing the orientation on all its edges, and avoiding repetition. For an undirected graph $G$, $top(G) = G$. An example of a directed graph is represented in Figure 2(a) with its topology in Figure 2(b).

*Definition 3 (Two Hop Neighbor):* In the undirected graph $top(G) = (V, A')$, $k \in V$ is a two hop neighbor of $i \in V$, if there is a $j \in V$ such that $(i, j) \in A'$ and $(j, k) \in A'$.

*Definition 4 (Children, Parents and Kins):* In a directed graph $G = (V, A)$, for a node $j \in V$, the children of $j$ are defined as $\mathcal{C}_G(j) := \{i | (i, j) \in A\}$ and the parents of $j$ as $\mathcal{P}_G(j) := \{i | (j, i) \in A\}$. Kins of $j \in V$ are defined as, $\mathcal{K}_G(j) := \{i | i \neq j \text{ and } i \in \mathcal{C}_G(j) \cup \cup \mathcal{P}_G(j) \cup \mathcal{P}_G(\mathcal{C}_G(j))\}$.

Note that the kin relation is a symmetric relationship; $i \in \mathcal{K}_G(j)$ if and only if $j \in \mathcal{K}_G(i)$.

*Definition 5 (Kin-graph):* Given an oriented graph $G = (V, A)$, its kin-graph is the undirected graph $\tilde{G} = (V, \tilde{A})$, where
$$\tilde{A} := \{(i, j) | i \in \mathcal{K}_G(j), j \in V\},$$
and is denoted as $kin(G) := \tilde{G}$.

*Definition 6:* (Path) A path is a non empty undirected graph $P = (V, E)$ with $V = \{x_0, x_1, \cdots, x_k\}$ and $E = \{(x_0, x_1), (x_1, x_2), \cdots, (x_{k-1}, x_k)\}$. We will denote a path by $x_0 - x_1 - x_2 - \cdots - x_{k-1} - x_k$.

*Definition 7:* (Cycle) A path $P$ of length at least 2 with the edge set $\{(x_0, x_1), (x_1, x_2), \cdots, (x_{k-1}, x_k), (x_k, x_0)\}$ is a cycle.

*Definition 8:* (Connectedness) A non-empty undirected graph $G := (V, A)$ is connected if for any $a, c \in V$ there exists a path of the form $a - b_1 - b_2 - \cdots - b_m - c$ in $G$.

*Definition 9:* (Tree) A connected undirected graph without cycles is called a tree. There is a unique path between any two nodes in a tree.

*Definition 10:* (Leaf Node/ Non leaf Node of a Tree) In a tree $\mathcal{T}$, a node with degree 1 is called a leaf node. Nodes with degree greater than 1 are called non leaf nodes.

## III. LINEAR DYNAMIC GRAPHS

In this section we introduce a class of models for the description of a network of JWSS (Jointly Wide Sense Stationary) processes. The node dynamics of the $j-th$ node is given by:
$$x_j = e_j + \sum_{i=1, i \neq j}^{m} \mathcal{H}_{ji}(z) x_i, j = 1, 2, \cdots, m.$$

If an agent $i$ 'influences' another agent $j$, that is, $\mathcal{H}_{ji}(z) \neq 0$, a directed edge is drawn from $i$ to $j$ and a directed graph is obtained. The noise component at each node $e_j$ is assumed to be zero mean WSS and uncorrelated with $\{e_k\}_{k=1, k \neq j}^{m}$.

*Definition 11 (Linear Dynamic Graph [15]):* A Linear Dynamic Graph $\mathcal{G}$ is defined as a pair $(\mathbb{H}(z), E)$ where
- $E = (e_1 \ldots e_m)'$ is a vector of $m$ uncorrelated WSS processes $\{e_j\}_{j=1}^{m}$. Thus, the power spectral density matrix of $E$, $\Phi_E(z)$ is a diagonal matrix of size $m \times m$.
- $\mathbb{H}(z)$ is a $m \times m$ matrix of stable transfer functions in $\mathcal{F}$ such that diagonal entries $\mathbb{H}(j, j)(z) = \mathbf{0}$, for $j = 1, ..., m$ and $\mathbb{H}(j, i)(z) = \mathcal{H}_{ji}(z), i \neq j$ is the $(j, i)$ entry of $\mathbb{H}(z)$.

A compact expression for the output processes $\{x_j\}_{j=1}^{m}$ of the LDG is,
$$X(k) = \mathbb{H}(z) X(k) + E(k), \tag{1}$$

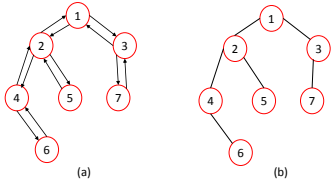

Fig. 3. (a) A bi-directed LDG and (b) its associated undirected graph.

where $X = [x_1, \cdots, x_m]'$. Let $V := \{x_1, ..., x_m\}$ and let $A := \{(x_j, x_i) | \mathcal{H}_{ji}(z) \neq 0\}$. The pair $G = (V, A)$ is the associated directed graph of the LDG. Nodes and edges of a LDG refer to the nodes and edges of the graph $G$.

A LDG $(\mathbb{H}(z), E)$ is well-posed if each entry of $(\mathbb{I} - \mathbb{H}(z))^{-1}$ is stable and topologically detectable if $\Phi_{e_j}(e^{\iota\omega}) > 0$ for any $\omega \in [-\pi, \pi]$ and $j = 1, ..., m$. In this article we focus on a restricted class of LDGs; bi-directed LDGs, which is defined below.

*Definition 12 (bi-directed LDG):* A LDG $(\mathbb{H}(z), E)$ whose associated graph is bidirectional is called a bi-directed LDG. Note that in an bi-directed LDG $\mathcal{H}_{ji}(z) \neq 0$ almost surely implies that $\mathcal{H}_{ij}(z) \neq 0$ almost surely. The associated graph of a bi-directed LDG could be interpreted as an undirected graph (see Figure 3 (b)).

## IV. LEARNING KIN GRAPH FROM DATA USING WIENER FILTERING

In this section we present Algorithm 1, which recovers the kin-graph of the underlying LDG from the observed data using multivariate Wiener filtering [15]. The multivariate Wiener filter of estimating $x_j$ from $x_{\bar{j}} := \{x_1, ..., x_{j-1}, x_{j+1}, ..., x_m\}$ is given by, $W_j(z) = [W_{j1}(z) \; ... \; W_{jj-1}(z) \; W_{jj+1}(z) \; \cdots W_{jm}(z)] = \Phi_{x_j x_{\bar{j}}}(z) \Phi_{x_{\bar{j}} x_{\bar{j}}}^{-1}(z)$.

---

**Algorithm 1** Topology Learning using Wiener Filtering

**Input:** Time series $x_i$ for nodes $i \in \{1, 2, ..., m\}$
**Output:** The kin graph of $\mathcal{T}$, $\mathcal{T}' = (V, \mathcal{E}_{\mathcal{T}'})$.

1: Edge set $\mathcal{E}_{\mathcal{T}'} \leftarrow \{\}$
2: **for all** $j \in \{1, 2, ..., m\}$ **do**
3:     Compute $W_j(z)$
4:     **for all** $i \in \{1, 2, ..., m\}, i \neq j$ **do**
5:         **if** $W_{ji}(z) \neq 0$ **then**
6:             $\mathcal{E}_{\mathcal{T}'} \leftarrow \mathcal{E}_{\mathcal{T}'} \cup \{(i, j)\}$
7:         **end if**
8:     **end for**
9: **end for**

---

*Remark 1:* It is proven in [15] that if $W_{ji}(z) \neq 0$ then $i$ and $j$ are kins. The converse is also true except for pathological cases (see [15]). Thus, we assume our LDG transfer functions are not from the pathological set, thereby enabling Algorithm 1 to recover the kin graph of the LDG. The kin graph obtained after application of Algorithm 1 is an undirected graph, hence, direction of the link cannot be inferred from the output of Algorithm 1. Note that Algorithm 1 is driven by the power spectral density matrices, which can be computed solely from the measured data.

Consider a bi-directed LDG $(\mathbb{H}(z), E)$ with the associated undirected graph $G = (V, A)$. Let the output of the LDG be given by $X = (x_1, ..., x_m)'$. The Wiener filtering based topology reconstruction leads to an undirected graph $G' = (V, A')$ such that $A' = A \cup$ {edges between two hop neighbors in $G$}.

## V. NETWORK RECONSTRUCTION FOR BI-DIRECTED LDGS WITH TREE TOPOLOGY

In this section we restrict our to attention to bi-directed LDGs whose underlying topology is a tree $\mathcal{T}$. Applying Algorithm 1 described in the previous section to a bi-directed LDG with a tree topology $\mathcal{T} := (V, \mathcal{E}_{\mathcal{T}})$ results in the topology $\mathcal{T}'$ with edge set $\mathcal{E}_{\mathcal{T}'} = \mathcal{E}_{\mathcal{T}} \cup \{(x_i, x_j) | x_i, x_j \in V$ are two hop neighbors in $\mathcal{T}\}$. Note that $\mathcal{E}_T \subset \mathcal{E}_{T'}$. We assume that there exist a path of length at least four in $\mathcal{T}$ and present analytical results leading to algorithms that eliminate the spurious two hop neighbor links in $\mathcal{T}'$. Thus, the generative tree topology $\mathcal{T}$ is exactly determined. For the subsequent developments we will need the notion of d-separation, which is defined below.

*Definition 13 (d-separation [10]):* In a directed graph $(V, A)$, $I, J, Z$ be three disjoint subsets of $V$. Then $dsep(I, Z, J)$ (to be read as $Z$, d-separates $I$ and $J$) if and only if every path $p$ from a node $i$ in $I$ to a node $j$ in $J$ satisfies the following conditions

- $p$ contains a chain of the form $\pi_1 \to z \to \pi_2$ or a fork of the form $\pi_1 \leftarrow z \to \pi_2$ such that $z$ belongs to the set $Z$,
- $p$ contains a collider of the form $\pi_1 \to x_c \leftarrow \pi_2$ such that $x_c$ or its descendants do not belong to the set $Z$,

where, $\pi_1$ and $\pi_2$ are nodes in $V$.

*Theorem 5.1:* Consider a directed graph $(V, A)$ such that $I, J, Z$ are disjoint and form a partition of $V$. Then $dsep(I, Z, J)$ if and only if there is no edge of the form

1) $i \to j$,
2) $j \to i$,
3) $i \to z$ and $z \leftarrow j$,

where, $i, j$ and $z$ are nodes in $I, J$ and $Z$ respectively.

*Proof:* See [28] for the proof. ∎

*Theorem 5.2:* In the bi-directed LDG $(\mathbb{H}(z), E)$ with a tree topology $\mathcal{T}$ and non leaf nodes $a, b$, $a - b$ is an edge in $\mathcal{T}$ if and only if there exist nodes $c$ and $d$ distinct from $a$ and $b$ such that $dsep(c, \{a, b\}, d)$ holds.

*Proof:* ($\Rightarrow$) It is given that $a \rightleftharpoons b$ is a link between non leaf nodes $a$ and $b$ in the bi-directed LDG associated with the tree $\mathcal{T}$. As $a$ and $b$ are non leaf nodes, there exist nodes $c$ and $d$ on opposite sides of the edge $a \rightleftharpoons b$ such that $c \rightleftharpoons a \rightleftharpoons b \rightleftharpoons d$, is a part of the LDG as shown in Figure 4 (a), where, one can show that $A \cup B \cup C \cup D \cup \{a, b, c, d\} = V$ and that there exist no path between any two sets amongst $A, B, C$ and $D$ in $\mathcal{T}$ that does not involve nodes $a$ or $b$. Consider $I := A \cup C \cup c$, $J := B \cup D \cup d$ and $Z := \{a, b\}$. Thus, applying Theorem 5.1 with the partition $I, J, Z$ of $V$ as defined above, we conclude that $dsep(I, Z, J)$ holds, which implies, $dsep(c, \{a, b\}, d)$. This completes the proof in the forward direction.

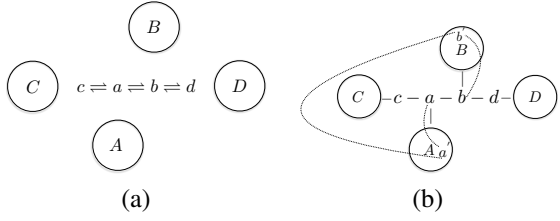

Fig. 4. (a) The bi-directed LDG associated with $\mathcal{T}$ with the true link $a \rightleftharpoons b$ and nodes $c, d$, and, (b) its topology $\mathcal{T}$ with a hypothetical link between sets $A$ and $B$, as described in the proof below.

($\Leftarrow$) The proof is by contradiction and left to the reader. ∎

*Theorem 5.3:* If $a - b$ is a edge in $\mathcal{T}$ such that $a$ is a leaf node, then there exist no nodes $c, d$ distinct from $a$ and $b$ such that $dsep(c, \{a, b\}, d)$ holds.

*Proof:* The proof uses Theorem 5.1, the property of a tree topology that there is a unique path between every pair of nodes and is left to the reader. ∎

*Definition 14 (Moralized Graph):* The graph obtained by connecting all parents having a common child in a directed graph with an undirected edge and then replacing all directed edges with undirected edges is its moralized graph.

*Remark 2:* The kin graph obtained using Wiener reconstruction(see Algorithm 1 and Remark 1) is the moralized graph of the directed graph associated with the LDG.

*Definition 15 (Ancestors):* In a directed graph $G = (V, A)$, node $j \in V$ is an ancestor of node $k \in V$ if there is a directed sub-graph of $G$ of the form $j \to \pi_1 \to \cdots \to \pi_l \to k$ from $j$ to $k$, where, $\{\pi_1, \cdots, \pi_l\} \in V$.

*Definition 16 (Ancestral Set):* In a directed graph $(V, A)$, the ancestral set of a node $j$ is the collection of ancestors of $j$ including $j$ itself and is denoted by $an(j)$. Given a collection of nodes $B$ subset of $V$, the ancestral set of $B$, $an(B) := \bigcup_{j \in B} an(j)$.

*Definition 17 (Ancestral Graph):* Given a directed graph $G = (V, A)$, the ancestral graph of a set of nodes $B$ subset of $V$ is the graph $G_{an(B)} = (an(B), E(an(B)))$ obtained from $G$ by removing all nodes not in $an(B)$.

*Theorem 5.4:* Let $I, J, Z$ be three disjoint sets of nodes in a directed graph $G = (V, A)$ that can have directed cycles. Then $I, J$ are d-separated by $Z$ in $G$ if and only if they are separated by $Z$ in the moral ancestral graph of $I, J, Z$, that is, $dsep(I, Z, J)$ in $G$ if and only if $sep(I, Z, J)$ in the moralized graph of $G_{an(I \cup Z \cup J)}$, where, $sep(I, Z, J)$ means after removing the nodes $Z$ there is no path between any node in $I$ and any node in $J$.

*Proof:* See [29], [30] for the proof. ∎

*Corollary 5.1:* Consider a bi-directed LDG $(\mathbb{H}(z), E)$ with a topology of a tree $\mathcal{T}$ whose vetrex set be $V$. If $I, J, Z$ are disjoint and form a partition of $V$, then $sep(I, Z, J)$ in the kin graph $\mathcal{T}'$ implies $dsep(I, Z, J)$ in the bi-directed LDG associated with $\mathcal{T}$.

*Proof:* The proof is left to the reader. ∎

Based on Theorem 5.2 and Theorem 5.3, we present Algorithm 2 which identifies the set of non leaf nodes $V_{nl}$, the set of leaf nodes $V_l$ and eliminates the spurious edges between non leaf node pairs and leaf node pairs to obtain the graph $\overline{\mathcal{T}} := (V, \mathcal{E}_{\overline{\mathcal{T}}})$. Set $\mathcal{E}_{\overline{\mathcal{T}}}$ is the edge set obtained after removing each spurious edge between two non-leaf nodes or two leaf nodes from $\mathcal{E}_{\mathcal{T}'}$.

**Algorithm 2** Elimination of spurious edges involving non leaf nodes in kin graph $\mathcal{T}'$

**Input:** $\mathcal{T}' = (V, \mathcal{E}_{\mathcal{T}'})$
**Output:** $\overline{\mathcal{T}} = (V, \mathcal{E}_{\overline{\mathcal{T}}})$

1: Edge set $\mathcal{E}_{\overline{\mathcal{T}}} \leftarrow \{\}$
2: **for all** edge $a - b$ in $\mathcal{E}_{\mathcal{T}'}$ **do**
3:     **if** $Z := \{a, b\}$ there exist $I \neq \{\phi\}$ and $J \neq \{\phi\}$ such that $sep(I, Z, J)$ holds in $\mathcal{T}'$ **then**
4:         $V_{nl} \leftarrow V_{nl} \cup \{a, b\}, \mathcal{E}_{\overline{\mathcal{T}}} \leftarrow \mathcal{E}_{\overline{\mathcal{T}}} \cup \{(a, b)\}$
5:     **end if**
6: **end for**
7: $V_l \leftarrow V - V_{nl}$
8: **for all** $a \in V_l, b \in Vnl$ with $(a, b) \in \mathcal{E}_{\mathcal{T}'}$ **do**
9:     $\mathcal{E}_{\overline{\mathcal{T}}} \leftarrow \mathcal{E}_{\overline{\mathcal{T}}} \cup \{(a, b)\}$
10: **end for**

Since the underlying topology is a tree, so each leaf node is connected to only one non leaf node (and no other leaf node) in $\mathcal{T}$. Note that the only spurious edges that exist in $\overline{\mathcal{T}}$ connect a leaf node (set $V_l$) with a non-leaf nodes (in set $V_{nl}$). Next we focus our attention on elimination of such spurious edges between a leaf node and a non-leaf node. Consider leaf node $x_i \in V_l$. Note that the neighbors of $x_i$ in $\overline{\mathcal{T}}$ comprise of non-leaf nodes of its kin-set in $\mathcal{T}$. Wlog assume that the kin set of node $i$ in $\mathcal{T}$ excluding the other leaf nodes is denoted by, $K_{\mathcal{T}}(x_i) = \{x_{j_k}\}_{k=1, j_k \neq i}^m$, where each $x_{j_k} \in V_{nl}$.

*Theorem 5.5:* Suppose $x_i$ is a leaf node in $\mathcal{T}$ such that $card(K_{\mathcal{T}}(x_i)) = 2$, that is, $K_{\mathcal{T}}(x_i) = \{x_{j_1}, x_{j_2}\}$. If $x_i - x_{j_1}$ is the edge in $\mathcal{T}$ then $x_{j_2}$ is the only non leaf two hop neighbor of $x_i$ in the graph $\overline{\mathcal{T}}$ with the $x_i - x_{j_2}$ edge removed.

*Proof:* The proof uses the idea that the spurious link would provide new two hop neighbors to arrive at a contradiction and is left to the reader. ∎

*Theorem 5.6:* Suppose $card(K_{\mathcal{T}}(x_i)) \geq 3$, then $x_{j_k}$ is the only common non leaf neighbor of each node in $K_{\mathcal{T}}(x_i) \cup \{x_i\} \backslash \{x_{j_k}\}$ in $\overline{\mathcal{T}}$, if and only if, $x_i - x_{j_k}$ is the edge in $\mathcal{T}$ and all other edges connecting $x_i$ to nodes in $K_{\mathcal{T}}(x_i) \backslash x_{j_k}$ are spurious(that is they are not in $\mathcal{T}$).

*Proof:* The proof is based on contradiction and is left to the reader. ∎

Thus, by analyzing the adjacency matrix of the reconstructed topology $\overline{\mathcal{T}}$, spurious links associated with leaf nodes can be eliminated from the set $\overline{\mathcal{T}}$ to recover the edge set associated $\mathcal{T}^*$. In the case of persistently exciting time series data and infinite samples, $\mathcal{T}^* = \mathcal{T}$. The steps to obtain the estimate $\mathcal{T}^*$ of the true topology $\mathcal{T}$ is described in Algorithm 3.

In this section we presented two algorithms, to be applied in the order Algorithm 2 followed by Algorithm 3 which build on the kin graph obtained from Algorithm 1 to recover the exact topology of a bi-directed LDG whose associated

**Algorithm 3** Elimination of spurious edges involving leaf nodes from $\overline{\mathcal{T}}$ to obtain the estimate $\mathcal{T}^*$

**Input:** $\overline{\mathcal{T}} = (V, \mathcal{E}_{\overline{\mathcal{T}}}), V_l, V_{nl}$
**Output:** $\mathcal{T}^* = (V, \mathcal{E}_{\mathcal{T}^*})$

1: Edge set $\mathcal{E}_{\mathcal{T}} \leftarrow \{\}$
2: **for all** $a \in V_l$ **do**
3:    **if** $card(K_{\mathcal{T}}(a)) \geq 3$ **then**
4:       determine $b_j \in V_{nl}$ in $K_{\mathcal{T}}(a)$ which is a neighbor of all nodes in $K_{\mathcal{T}}(a) \cup \{a\}$, and,
5:       $\mathcal{E}_{\overline{\mathcal{T}}} \leftarrow \mathcal{E}_{\overline{\mathcal{T}}} - \{(a, b_i)\}_{i \neq j, b_i \in K_{\mathcal{T}}(a)}$
6:    **end if**
7:    **if** $card(K_{\mathcal{T}}(a)) = 2$ **then**
8:       Use Theorem 5.5 to determine $x_{j_2}$ and update $\mathcal{E}_{\overline{\mathcal{T}}} \leftarrow \mathcal{E}_{\overline{\mathcal{T}}} - (a, x_{j_2})$
9:    **end if**
10: **end for**
11: $\mathcal{E}_{\mathcal{T}^*} \leftarrow \mathcal{E}_{\overline{\mathcal{T}}}$

graph is a tree. In the next section we present a brief description of the dynamics of a network of generators to motivate examples from the grid for application of the three stage reconstruction procedure.

## VI. APPLICATIONS IN POWER SYSTEMS

In this section we present a model for a radial network of generators in a power system and demonstrate that this model fits the bi-directed LDG framework. The topology of the power network is modeled by a graph $G = (V, A)$ where nodes in $V$ represent buses and edges in $A$ represent transmission lines. Each bus $i$ is associated with a voltage phase angle $\theta_i$ that determines its state. The state dynamics in each node in the system is governed by the following second order equation:

$$m_i \ddot{\theta}_i + d_i \dot{\theta}_i = -p_{e,i} + p_{in,i}, \ i = 1, 2, \cdots, m, \quad (2)$$

Here $p_{in,i}$ and $p_{e,i}$ are power input and electrical power output respectively at node $i$. Equation (2) describes the swing dynamics for $\theta_i$ with $m_i > 0$ and $d_i > 0$ being the rotational inertia and damping of the generator or rotating load device at node $i$ respectively. The electrical power output is complex valued (AC current). However, for small dynamics, we assume constant voltage magnitudes, purely inductive lines and a small signal approximation [31]. The electrical power output from node $i$ is then given by [5]:

$$p_{e,i} = \sum_{k=1}^{m} b_{ik}(\theta_i - \theta_k), i \in \{1, \cdots, m\}, \quad (3)$$

where $b_{i,k} \geq 0$ is the susceptance on the line between nodes $i, k \in A$. Indeed, $b_{ik}(\theta_i - \theta_k)$ represents the power flow from node $i$ to node $k$ in the network. Using $f_i = \dot{\theta}_i$, we can write the following linear state space representation of the power system dynamics

$$\begin{bmatrix} \dot{\theta} \\ \dot{f} \end{bmatrix} = \begin{bmatrix} \mathbf{0} & \mathbf{I} \\ -\mathbf{M}^{-1}\mathbf{L} & -\mathbf{M}^{-1}\mathbf{D} \end{bmatrix} \begin{bmatrix} \theta \\ f \end{bmatrix} + \begin{bmatrix} \mathbf{0} \\ \mathbf{M}^{-1} \end{bmatrix} p_{in}, \quad (4)$$

where, $\mathbf{M} = \text{diag}\{m_i\}, \mathbf{D} = \text{diag}\{d_i\}$ are diagonal mass and damping matrices of the network, $\mathbf{I} \in \mathbb{R}^{m \times m}$ is the identity matrix, $\mathbf{0} \in \mathbb{R}^{m \times m}$ is the zero matrix. $\mathbf{L} = \mathbf{L}^T \in \mathbb{R}^{m \times m}$ is the network susceptance weighted Laplacian with off-diagonal elements $\mathbf{L}(i,k) = -b_{ik}$ and diagonal elements $\mathbf{L}(i,i) = \sum_{k=1, k \neq i}^{m} b_{ik}$. The state vector $[\theta^T \ f^T]^T \in \mathbb{R}^{2m}$ is comprised of a stacked vector of all angles $\theta \in \mathbb{R}^m$ and frequencies $f \in \mathbb{R}^m$. The vector $p_{in} \in \mathbb{R}^m$ is a collection of power inputs to all nodes in the system. The output equation of the network in discrete time is given by,

$$X(k) = \begin{bmatrix} \mathbf{I} & \mathbf{0} \end{bmatrix} \begin{bmatrix} \theta(k) \\ \omega(k) \end{bmatrix}, \quad (5)$$

where the states $\theta$ is measured using nodal PMUs [21]. All state and input variables in (4) can be interpreted as deviations from the steady state, $p_{in}$ thus represents ambient deviations from input and modelled as a vector of zero mean uncorrelated WSS processes.

The discretized version of the continuous time network dynamics characterized by (4) and (5) takes the form $X(k) = \mathbb{H}(z)X(k) + E$, which justifies the modeling of a network of generators as a LDG $(\mathbb{H}(z), E)$. The bi-directedness arises due to the symmetric nature of $\mathbf{L}$. In the next section we apply the three stage topology identification process to $LDGs$ of generators governed by the swing equations with the underlying topology being a tree.

## VII. RESULTS

We apply the algorithms described in the previous sections to a network of generators whose underlying topology is a tree $\mathcal{T}$. Consider the five node chain of generators as shown in Figure 5(I)(a). The dynamics of the chain is simulated using (2) and the input $p_{in}$ is considered to be zero mean correlated WSS process. Applying the Wiener filtering based network reconstruction procedure (Algorithm 1) on the output data from the chain, results in the kin graph of the chain $\mathcal{T}'$ as shown in Figure 5(I)(b). Next we apply Algorithm 2 on the kin graph $\mathcal{T}'$ and obtain $V_{nl} = \{2, 3, 4\}$, $\mathcal{E}_{\mathcal{T}} = \{(2,3),(3,4)\}$ and $V_l = \{1, 5\}$. The reduced graph $\overline{\mathcal{T}}$ is shown in Figure 5(I)(c). Applying Algorithm 3 on $\overline{\mathcal{T}}$ eliminates the spurious links involving leaf nodes and leads to $\mathcal{E}_{\mathcal{T}^*} = \{(1,2),(2,3),(3,4),(4,5)\} = \mathcal{E}_{\mathcal{T}}$ as shown in Figure 5(I)(d).

Next, we apply our three stage topology identification procedure to the tree shown in Figure 5(II) which is derived from the IEEE 39 bus [32], [33] by removing the cycles (four lines removed). There are 11 generator nodes (shown in yellow) and the rest are load nodes. The simulation parameters are available in [33]. We apply the three stage reconstruction procedure on the simulated phase angle information ($10^7$ samples per node) to recover the topology shown in Figure 5. The reconstruction is exact and is not shown separately.

## VIII. CONCLUSIONS

We develop a Wiener filter based three-stage algorithm to reconstruct the exact topology of a bidirectional radial network of dynamically related WSS processes. The advantage of our approach is that it does not use any prior information about the network except the fact that the underlying topology is a tree. In particular, it is extremely useful in estimating the operating topology of radial power grid using phase angle measurements of the swing equations. We illustrate the proposed algorithm with an example of a 5 node chain and also on a modified tree-version of the IEEE 39 bus test system.

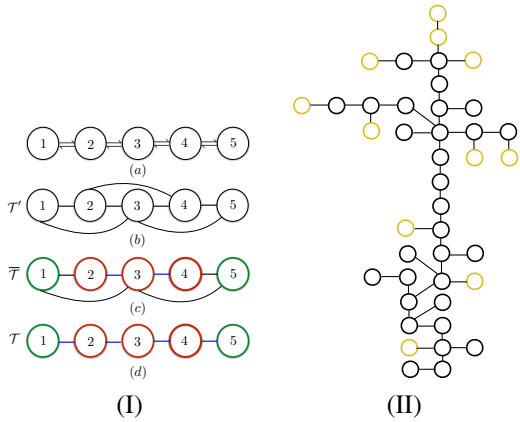

Fig. 5. (I) (a) Chain of 5 generators governed by Swing equations (2), (b) the kin graph $\mathcal{T}'$ of the chain obtained using Algorithm 1(Wiener filtering), (c) inference of nodes $2, 3, 4$ as non leaf nodes(red), nodes $1, 5$ as leaf nodes(green) and the edges $(2, 3), (3, 4)$ as edges between non leaf nodes(blue) to obtain $\overline{\mathcal{T}}$ by using Algorithm 2 on $\mathcal{T}'$, (d) elimination of spurious edges associated with leaf nodes 1 and 5(black edges in (c)) to obtain $\mathcal{T}$ by using Algorithm 3 on $\overline{\mathcal{T}}$, (II) IEEE 39 bus network topology with the loops removed.

IX. ACKNOWLEDGMENTS

The authors S. Talukdar, D. Materassi and M. V. Salapaka acknowledge the support of ARPA-E for supporting this research through the project titled 'A Robust Distributed Framework for Flexible Power Grids' via grant no. DE-AR000071. D. Deka acknowledges the support of funding from the U.S. Department of Energy's Office of Electricity as part of the DOE Grid Modernization Initiative.

REFERENCES

[1] E. Bullmore and O. Sporns, "Complex brain networks: graph theoretical analysis of structural and functional systems," *Nature Reviews Neuroscience*, vol. 10, no. 3, pp. 186–198, 2009.
[2] K. Knorr Cetina and A. Preda, *The sociology of financial markets*. Oxford University Press, 2004.
[3] J.-F. Rual, K. Venkatesan, T. Hao, T. Hirozane-Kishikawa, A. Dricot, N. Li, G. F. Berriz, F. D. Gibbons, M. Dreze, N. Ayivi-Guedehoussou et al., "Towards a proteome-scale map of the human protein–protein interaction network," *Nature*, vol. 437, no. 7062, pp. 1173–1178, 2005.
[4] J. F. Donges, Y. Zou, N. Marwan, and J. Kurths, "Complex networks in climate dynamics," *The European Physical Journal Special Topics*, vol. 174, no. 1, pp. 157–179, 2009.
[5] P. Kundur, N. J. Balu, and M. G. Lauby, *Power system stability and control*. McGraw-hill New York, 1994, vol. 7.
[6] F. Bullo, J. Cortes, and S. Martinez, *Distributed control of robotic networks: a mathematical approach to motion coordination algorithms*. Princeton University Press, 2009.
[7] P. DâĂŹhaeseleer, S. Liang, and R. Somogyi, "Genetic network inference: from co-expression clustering to reverse engineering," *Bioinformatics*, vol. 16, no. 8, pp. 707–726, 2000.
[8] J. Chen and A. H. Sayed, "Diffusion adaptation strategies for distributed optimization and learning over networks," *IEEE Transactions on Signal Processing*, vol. 60, no. 8, pp. 4289–4305, 2012.
[9] J. Pearl, *Probabilistic reasoning in intelligent systems: networks of plausible inference*. Morgan Kaufmann, 2014.
[10] ——, *Causality*. Cambridge university press, 2009.
[11] M. I. Jordan, *Learning in graphical models*. Springer Science & Business Media, 1998, vol. 89.
[12] S. L. Lauritzen, *Graphical models*. Clarendon Press, 1996, vol. 17.
[13] V. Chandrasekaran, P. A. Parrilo, and A. S. Willsky, "Latent variable graphical model selection via convex optimization," *Ann. Statist.*, vol. 40, no. 4, pp. 1935–1967, 08 2012. [Online]. Available: http://dx.doi.org/10.1214/11-AOS949
[14] R. Dahlhaus, "Graphical interaction models for multivariate time series1," *Metrika*, vol. 51, no. 2, pp. 157–172, 2000.
[15] D. Materassi and M. V. Salapaka, "On the problem of reconstructing an unknown topology via locality properties of the wiener filter," *IEEE Transactions on Automatic Control*, vol. 57, no. 7, pp. 1765–1777, 2012.
[16] S. Talukdar, M. Prakash, D. Materassi, and M. V. Salapaka, "Reconstruction of networks of cyclostationary processes," in *IEEE 54th Annual Conference on Decision and Control (CDC), 2015*. IEEE, 2015, pp. 783–788.
[17] J. Etesami and N. Kiyavash, "Directed information graphs: A generalization of linear dynamical graphs," in *2014 American Control Conference*. IEEE, 2014, pp. 2563–2568.
[18] D. Deka, S. Backhaus, and M. Chertkov, "Estimating distribution grid topologies: A graphical learning based approach," in *Power Systems Computation Conference (PSCC)*, 2016.
[19] R. Hoffman, "Practical state estimation for electric distribution networks," in *IEEE PES Power Systems Conference and Exposition*. IEEE, 2006, pp. 510–517.
[20] D. Deka, S. Backhaus, and M. Chertkov, "Structure learning and statistical estimation in distribution networks - part i," *arXiv preprint arXiv:1501.04131*, 2015.
[21] A. Phadke, "Synchronized phasor measurements in power systems," *IEEE Computer Applications in Power*, vol. 6, no. 2, pp. 10–15, 1993.
[22] A. von Meier, D. Culler, A. McEachern, and R. Arghandeh, "Micro-synchrophasors for distribution systems," pp. 1–5, 2014.
[23] S. Bolognani, N. Bof, D. Michelotti, R. Muraro, and L. Schenato, "Identification of power distribution network topology via voltage correlation analysis," in *IEEE Decision and Control (CDC)*. IEEE, 2013, pp. 1659–1664.
[24] D. Deka, S. Backhaus, and M. Chertkov, "Learning topology of the power distribution grid with and without missing data," in *European Control Conference (ECC)*, 2016.
[25] ——, "Learning topology of distribution grids using only terminal node measurements," in *IEEE Smartgridcomm*, 2016.
[26] R. Sevlian and R. Rajagopal, "Feeder topology identification," *arXiv preprint arXiv:1503.07224*, 2015.
[27] R. Diestel, *Graph Theory*. Berlin, Germany: Springer-Verlag, 2006.
[28] J. T. Koster, "On the validity of the markov interpretation of path diagrams of gaussian structural equations systems with correlated errors," *Scandinavian Journal of Statistics*, vol. 26, no. 3, pp. 413–431, 1999.
[29] E. Castillo, J. M. Gutierrez, and A. S. Hadi, *Expert systems and probabilistic network models*. Springer Science & Business Media, 2012.
[30] J. Pearl and R. Dechter, "Identifying independencies in causal graphs with feedback," in *Proceedings of the Twelfth international conference on Uncertainty in artificial intelligence*. Morgan Kaufmann Publishers Inc., 1996, pp. 420–426.
[31] A. Abur and A. G. Exposito, *Power system state estimation: theory and implementation*. CRC press, 2004.
[32] T. Athay, R. Podmore, and S. Virmani, "A practical method for the direct analysis of transient stability," *IEEE Transactions on Power Apparatus and Systems*, no. 2, pp. 573–584, 1979.
[33] A. Pai, *Energy function analysis for power system stability*. Springer Science & Business Media, 2012.